# Efficient Continual Pre-training of LLMs for Low-resource Languages

**Arijit Nag**
IIT Kharagpur
arijitnag@iitkgp.ac.in

**Animesh Mukherjee**
IIT Kharagpur
animeshm@cse.iitkgp.ac.in

**Niloy Ganguly**
IIT Kharagpur
niloy@cse.iitkgp.ac.in

**Soumen Chakrabarti**
IIT Bombay
soumen@cse.iitb.ac.in

## Abstract

Open-source Large Language models (Os-LLMs) propel the democratization of natural language research by giving the flexibility to augment or update model parameters for performance improvement. Nevertheless, like proprietary LLMs, Os-LLMs offer poorer performance on low-resource languages (LRLs) than high-resource languages (HRLs), owing to smaller amounts of training data and underrepresented vocabulary. On the other hand, continual pre-training (CPT) with large amounts of language-specific data is a costly proposition in terms of data acquisition and computational resources. Our goal is to drastically reduce CPT cost. To that end, we first develop a new algorithm to select a subset of texts from a larger corpus. We show the effectiveness of our technique using very little CPT data. In search of further improvement, we design a new algorithm to select tokens to include in the LLM vocabulary. We experiment with the recent Llama-3 model and nine Indian languages with diverse scripts and extent of resource availability. For evaluation, we use IndicGenBench, a generation task benchmark dataset for Indic languages. We experiment with various CPT corpora and augmented vocabulary size and offer insights across language families.

## 1 Introduction

Large language models (LLMs) like GPT-4 (OpenAI et al., 2023), ChatGPT, Llama-2 (Touvron et al., 2023), Llama-3 (Dubey et al., 2024), PaLM (Chowdhery et al., 2022), *inter alia*, are opening up new possibilities for low-resource languages (LRLs). Until recently, collecting sufficient labeled LRL data to finetune LLMs for classification and generation tasks used to be challenging. Today, LLMs give decent performance with zero/few-shot inference. Having said that, there is still a substantial performance gap between high-resource languages (HRLs) and LRLs for LLMs (Hendy et al., 2023; Jiao et al., 2023; Bang et al., 2023). This is due to the fact that LRLs like Indic languages are still under-represented by recent LLMs, as shown in Figure 1: Compared to English, the average number of tokens required to generate a LRL word by these LLMs is substantially higher. The inability to represent a word with a single token may lead to suboptimal learning of context thus potentially affecting LLM's performance for LRL tasks. A feasible way to overcome such shortcoming is to initiate continual pre-training (CPT), specifically with LRL text.

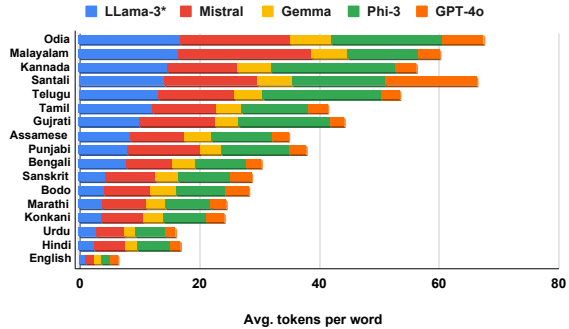

Figure 1: Average tokens generated per word for various Indic languages using different recent LLMs. The last column shows the performance in English.

CPT can help LLMs learn domains/languages that are un/under-explored in the pre-training stage. While this is a viable option to improve LLM's performance, training such gigantic models consumes expensive GPU resources and time, which makes it less feasible in resource-constrained setups. To address these issues and harness CPT's potential, we propose a two-pronged approach. First, we introduce a score-based method to select a small set of high-quality, language-specific training data. Concurrently, we implement a strategy to expand the token vocabulary in LLMs. This vocabulary augmentation improves the understanding of important words in low-resource languages, leading to further performance gains. The strategies and the rigorous experiments undertaken are detailed next.

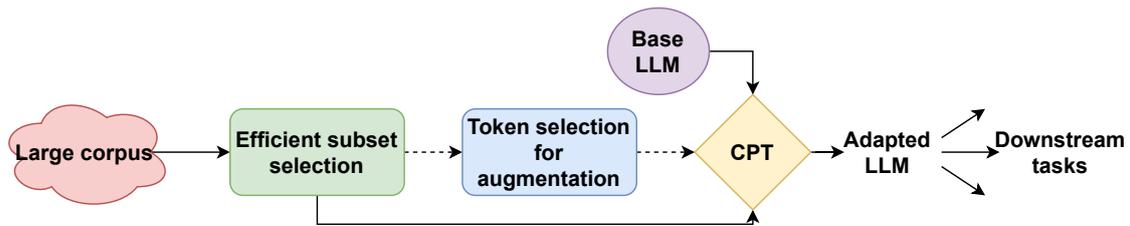

Figure 2: Sketch of proposed LLM CPT method. Dotted lines indicate optional steps.

**The two proposed methods**
**(1)** We propose a global + Local score for each sentence for selecting a small subset of text from an LRL training corpus to perform CPT and improve LLM performance. Experiments show significant performance boost.
**(2)** We propose a method to augment the token vocabulary of the LLM to further improve LRL task performance in certain situations.
**Experiments** We present a comprehensive study on the recently released and very popular whitebox LLM LLama-3-8B, applying our CPT methods to nine Indian languages in six scripts, covering three resource levels (High, Mid, Low) (Singh et al., 2024) over five LRL generation tasks provided by IndicGenBench (Singh et al., 2024), including summarization, machine translation, and question-answering.
**Observations** Single/limited word prediction tasks like QA are more sensitive to vocabulary augmentation compared to multi-word generation tasks like summarization or machine translation; the effect of vocabulary augmentation on tokenization varies across scripts; and larger CPT corpus and vocabulary do not always convert to performance improvements.

## 2 Related work

Language models use diverse subword tokenization algorithms like Byte-Pair Encoding (BPE) (Sennrich et al., 2016), SentencePiece (Kudo and Richardson, 2018), WordPiece (Schuster and Nakajima, 2012), and Unigram (Kudo, 2018). Due to the limited size of an LLM's token vocabulary, over-fragmentation (Muller et al., 2021; Rust et al., 2021; Ahia et al., 2023; Petrov et al., 2023) is a common problem, especially for multilingual models where not all languages get equal representation. Apart from task performance degradation (Hendy et al., 2023; Jiao et al., 2023; Bang et al., 2023; Toraman et al., 2023; Fujii et al., 2023), over-fragmentation leads to slow inference (Petrov et al., 2023; Hofmann et al., 2022) and increased training and inference/generation cost (Ahia et al., 2023; Petrov et al., 2023; Nag et al., 2024). Various mitigation methods have been proposed, including vocabulary expansion (Chau et al., 2020; Cui et al., 2024; Balachandran, 2023; Fujii et al., 2024; Yamaguchi et al., 2024a) and replacing existing tokens in the vocabulary with new ones (Minixhofer et al., 2022; Dobler and de Melo, 2023; Ostendorff and Rehm, 2023; Downey et al., 2023). In recent work such as ChineseLlama (Cui et al., 2024) and TamilLlama (Balachandran, 2023), the authors add new language-specific tokens and then pretrain the model with large amounts of training data. Contemporary LLMs are gigantic, consuming much computing resources and time to train, making these proposals less feasible for resource-constrained setups like academia or small research labs. More recently, Yamaguchi et al. (2024b) and Tejaswi et al. (2024) explore CPT of LLMs while varying the corpus, additional vocabulary and embedding initialization techniques. However, they do not focus on strategies to select corpus and vocabulary.

In contrast, in this work, we propose a global+local joint rank-based system to first select the small-scale training corpus and then augment the LLM's vocabulary with additional language-specific tokens for CPT. With a small amount of informative training data and added vocabulary, we show substantial LLM performance improvement for Indic languages.

## 3 Proposed method

In this work, we design a two-stage approach to improve LLM's performance with reduced resource requirements. In the first stage, we select a subset of the available LRL corpus, and in the next stage, we select prospective new tokens for vocabulary augmentation. These two together are used for the purpose of CPT as shown in Figure 2. As the fig-

**Algorithm 1** Corpus selection for CPT.

**Inputs:**
- Large training corpus $C_l$
- Number of sentences to select $K$
- LLM tokenizer $\mathcal{T}$
- Parameters for weighted average $\alpha, \beta$

1: $\mathcal{W} \leftarrow$ vocabulary from corpus $C_l$
2: fill $WC$ (word count dictionary) using $\mathcal{W}, C_l$
3: $SWC \leftarrow \{\}$ /* subword count dictionary */
4: **for** $w \in \mathcal{W}$ **do**
5:    **for** subword tokens $t \in \mathcal{T}(w)$ **do**
6:       $SWC[t] += WC[w]$
7: **for** each word $w$ **do**
8:    initialize $N[w] \leftarrow 0$
9: /* $N[w]$ will store aggregated popularity of subwords of $w$ relative to itself */
10: **for** $w \in \mathcal{W}$ **do**
11:    **for** $t \in \mathcal{T}(w)$ **do**
12:       $N[w] += SWC[t] - WC[w]$
13: fill $X_{co}$ with word-word co-occurrence matrix from $C_l$
14: /* co-occurrence within a context window */
15: $W_g \leftarrow$ PageRank($X_{co}$)
16: /* $W_g[w]$ stores the PageRank score of word $w$. */
17: initialize $R_l[s] \leftarrow 0$ for all sentences $s \in C_l$
18: /* Local sentence score table */
19: initialize $R_g[s] \leftarrow 0$ for all sentences $s \in C_l$
20: /* Global sentence score table */
21: initialize $R_j[s] \leftarrow 0$ for all sentences $s \in C_l$
22: /* Joint sentence score table */
23: **for** sentence $s \in C_l$ **do**
24:    **for** word $w \in s$ **do**
25:       $R_l[s] += N[w]$ /* popularity */
26:       $R_g[s] += W_g[w]$ /* importance */
27:    $R_j[s] = \alpha R_l[s] + \beta R_g[s]$
28: $C_r \leftarrow$ top-$K$ sentences by decreasing $R_j[s]$
29: **return** CPT training corpus $C_r$

ure shows, the second stage (token selection) is optional. Section 3.1 and 3.2 describe the corpus and vocabulary selection algorithms, respectively.

## 3.1 Stage I: Sentence selection

In this stage, the goal is to identify a subset of sentences from LRL corpus $C_l$ that will effectively enhance the LLM as a representative of the whole of $C_l$. We regard a sentence as a strong representative if it contains numerous 'important' words formed from popular subword tokens. These important words reflect the unique features of the corpus, while the popular tokens represent commonly used contexts.

**Popular subwords** In Algorithm 1, we first use the LLM's tokenizer to get all distinct subword tokens present in the corpus and compute their occurrence frequencies. Next, for a given word $w$ we compute the sum of the frequencies of its subwords. We now subtract the frequency of $w$ from this sum which indicates how much these tokens solely contribute to words other than $w$. If this difference is high then it implies that the subwords of $w$ contribute to many other words in the corpus and are thus more popular.

**Important words** From the LRL corpus, Algorithm 1 (line 13) also builds a graph where the words are nodes, and two words are connected if they co-occur in a predefined context window. For all of our experiments, we fixed the context window length as 5. The weighted adjacency matrix is $X_{co}$. Then we apply the PageRank algorithm (Page et al., 1999) on $X_{co}$ to get the PageRank score of each word in $W_g$. For a given sentence, we sum the PageRank values of the constituent words to assign a global score to the sentence (line 26). Note that global score $R_g[s]$ is LLM-agnostic.

Finally, we combine (global) importance and (local) popularity scores to obtain a weighted combination score for each sentence, and select the top sentences based on this final score.

## 3.2 Stage II: Vocabulary selection

Similar to sentence subset selection, we wish to find words from the selected sentences (output of Algorithm 1) that are contextually important and, at the same time, contain popular subwords which are shared by many words, making them vulnerable to distorted representation.

The initial parts of Algorithm 2 are identical to sentence selection, but then we create score maps $R_l[w], R_g[w], R_j[w]$ for *words* to be used to get prospective tokens for augmentation in the LLM vocabulary, not *sentences*. Here, we get the important words $V_{target}$ by sorting $w$ by decreasing $R_j$ values and choosing the words with top $Q$ percentile scores (line 21). In our experiments, we use the 50th percentile (median) as the threshold to avoid long tail words. Next, we create a dummy corpus $C_{dummy}$ by concatenating each word $w$ in $V_{target}$, $WC[w]$ number of times, separated by space (line 24). Finally, we pass the dummy corpus $C_{dummy}$, and the desired token size $K$ to a dictionary building and tokenization algorithm $\psi$ (line 26). For our case, we use the SentencePieceBPE tokenization algorithm.

To initialize the embedding of newly augmented tokens, we use the mean embedding of the constituent subwords generated by the existing tokenizer (Gee et al., 2022) and train them (update the embedding value) while doing CPT.

## 4 Experiment and results

To check the effectiveness of our two-stage CPT method, we use IndicGenBench (Singh et al.,

**Algorithm 2** Vocabulary extension before CPT.

**Inputs:**
- CPT corpus $C_{\text{CPT}}$
- (existing) LLM tokenizer $\mathcal{T}$
- Tokenizer training algorithm $\psi$
- Parameter for weighted average $\alpha, \beta$
- $Q$, top percentile of words to send to tokenizer
- $K$, the number of new tokens to include

1: $\mathcal{W} \leftarrow$ vocabulary from corpus $C_{\text{CPT}}$
2: fill $WC$ (word count dictionary) using $\mathcal{W}, C_{\text{CPT}}$
3: $SWC \leftarrow \{\}$                     /* subword count dictionary */
4: **for** word $w \in \mathcal{W}$ **do**
5:    **for** subword tokens $t \in \mathcal{T}(w)$ **do**
6:       $SWC[t] \mathrel{+}= WC[w]$
7: initialize $R_l[w] \leftarrow 0$ for all word $w \in \mathcal{W}$
8: /* Local word score table                                            */
9: Initialize $R_g[w] \leftarrow 0$ for all word $w \in \mathcal{W}$
10: /* Global word score table                                          */
11: Initialize $R_j[w] \leftarrow 0$ for all word $w \in \mathcal{W}$
12: /* Joint word score table                                           */
13: **for** word $w \in \mathcal{W}$ **do**
14:    **for** $t \in \mathcal{T}(w)$ **do**
15:       $R_l[w] \mathrel{+}= SWC[t] - WC[w]$
16: $X_{co} \leftarrow$ Word co-occurrence matrix of $C_l$
17: $R_g \leftarrow$ PageRank($X_{co}$)
18: **for** $w \in \mathcal{W}$ **do**
19:    $R_j[w] = \alpha\, R_l[w] + \beta\, R_g[w]$
20: $V_{target} \leftarrow \{\}$
21: sort $w$ by decreasing $R_j[w]$ and add top-$Q$ percentile words to $V_{target}$
22: $C_{dummy} \leftarrow$ empty string
23: /* Dummy corpus for training LLM tokenizer                          */
24: **for** word $w \in V_{target}$ **do**
25:    append $w$ to $C_{dummy}$ a total of $WC[w]$ times
26: $t_{aug} \leftarrow \psi(C_{dummy}, K)$
27: **return** $t_{aug}$, the tokens selected for augmentation

2024), a generation task benchmark dataset for Indic languages covering Cross-lingual Summarization, Machine Translation (MT) and Question-Answering (QA) tasks (see Figure 4 in the Appendix for dataset overview). For MT and QA tasks, there are two variants: one where the target language is one of the Indic languages (Flores(en→xx), XorQA(xx)), and the other where the target language is English (Flores(xx→en), XorQA(en)). For summarization, it is only from English to Indic languages (CrossSum). We experiment with *nine* Indic languages covering *six* (Devanagari, Bengali, Arabic, Telugu, Olchiki, Gurumukhi) different scripts and *three* (High/Mid/Low) types of resource availability as described in the existing work (Singh et al., 2024) and use the Llama-3-8B parameter model as our base LLM. We perform all our experiments in zero-shot setting both for off-the-shelf vanilla LLM and after doing the CPT over it. Details of LLM parameters and prompts are in the Appendix B (see Table 8 and Figure 5, respectively). For evaluation, we use Character-F1 (ChrF++ (Popović, 2017)) for Summarization and MT tasks and Token-F1 for QA tasks. To restrict the cost of experiments, we limit the CPT corpus size to 10K, 20K and 30K and, similarly, the augmented vocabulary size to 100, 200 and 300.

### 4.1 CPT corpus helps despite small size

In Table 1, we show the effect of CPT of the vanilla LLM with the small-sized ranked corpus that we obtain using Algorithm 1. We experiment with 10K, 20K and 30K top-ranked sentences as CPT corpus and report the best among them (denoted as **TR(Best)**). We use off-the-shelf vanilla Llama-3-8B model's performance as our baseline. We also report the change in performance (%) from vanilla to TR(Best) for individual languages as well as resource type availability. In general, we observe significant performance improvements for most of the tasks and languages. The improvements are progressively higher from the high-resource language group to the low-resource language group. This observation is expected as the vanilla LLMs are already well-trained in high-resource languages and may not get much benefit from CPT as compared to the resource-poor languages. Further for the QA tasks, both when the target language is Indic and English, we observe limited improvement for most of the cases and especially for English target (XorQA(en)) performance drops after CPT. This can be due to catastrophic forgetting of the English part as we do the CPT with Indic language-specific data and also as QA tasks performed here are limited word (1-2 words) prediction tasks, making it more vulnerable to such problems. In Section 4.5, we discuss a solution for them.

### 4.2 Sentence scoring and ranking help

To study the effect of corpus ranking we compare TR(Best) with BR(Best). We form 10K, 20K, 30K subsets with the least scoring sentences from the corpus, perform CPT and report the best performance among them as **BR(Best)**. In Table 1, we report the change in performance (%) from TR(Best) to BR(Best) for individual languages as well as based on resource type availability. We observe that TR(Best) outperforms BR(Best) across all tasks and languages except the QA tasks, showing the effectiveness of the ranking algorithm. It might be possible that top-ranked sentences lack diversity and may constrain the output token distribution. As QA tasks are sensitive to single-word prediction, it can affect performance adversely.

| Lang | Script | Type | Metric→<br>CPT data↓ | Target(xx)<br>Chrf++<br>CrossSum | Chrf++<br>Flores(en→xx) | Token-F1<br>XorQA(xx) | Target(en)<br>Chrf++<br>Flores(xx→en) | Token-F1<br>XorQA(en) |
|---|---|---|---|---|---|---|---|---|
| Urdu | Arabic | | Vanilla | 17.79 | 31.01 | 0.34 | 42.24 | 0.65 |
| | | | TR(Best) | 22.29 | 31.51 | 0.31 | 45.46 | 0.58 |
| | | | BR(Best) | 14.37 | 24.21 | 0.31 | 38.26 | 0.65 |
| | | | Vanilla→TR(↑) | 25.30% | 1.61% | -8.82% | 7.62% | -10.77% |
| | | | BR→TR(↑) | 55.11% | 30.15% | 0% | 18.82% | -10.77% |
| Bengali | Bengali | High | Vanilla | 16.09 | 28.45 | 0.61 | 41.41 | 0.64 |
| | | | TR(Best) | 17.35 | 28.97 | 0.63 | 43.42 | 0.58 |
| | | | BR(Best) | 14.69 | 24.44 | 0.67 | 44.81 | 0.66 |
| | | | Vanilla→TR(↑) | 7.83% | 1.83% | 3.28% | 4.85% | -9.38% |
| | | | BR→TR(↑) | 18.11% | 18.54% | -5.97% | -3.10% | -12.12% |
| Telugu | Telugu | | Vanilla | 13.21 | 25.59 | 0.28 | 39.65 | 0.61 |
| | | | TR(Best) | 16.51 | 25.57 | 0.37 | 39.31 | 0.59 |
| | | | BR(Best) | 14.39 | 23.34 | 0.33 | 39.53 | 0.67 |
| | | | Vanilla→TR(↑) | 24.98% | -0.08% | 32.14% | -0.86% | -3.28% |
| | | | BR→TR(↑) | 14.73% | 9.55% | 12.12% | -0.56% | -11.94% |
| | | | Avg(Vanilla→TR(↑)) | **19.37%** | **1.12%** | **8.87%** | **3.87%** | -7.81% |
| | | | Avg(BR→TR(↑)) | **29.32%** | **19.41%** | **2.05%** | **5.05%** | -11.61% |
| Sanskrit | Devanagari | | Vanilla | 7.69 | 12.35 | 0.43 | 30.35 | 0.55 |
| | | | TR(Best) | 13.63 | 15.15 | 0.31 | 33.71 | 0.42 |
| | | | BR(Best) | 12.57 | 16.21 | 0.41 | 31.47 | 0.39 |
| | | | Vanilla→TR(↑) | 77.24% | 22.67% | -27.91% | 11.07% | -23.64% |
| | | | BR→TR(↑) | 8.43% | -6.54% | -24.39% | 7.12% | 7.69% |
| Assamese | Bengali | Mid | Vanilla | 11.01 | 15.91 | 0.57 | 30.26 | 0.56 |
| | | | TR(Best) | 15.78 | 21.81 | 0.61 | 39.52 | 0.56 |
| | | | BR(Best) | 12.81 | 18.18 | 0.59 | 34.38 | 0.61 |
| | | | Vanilla→TR(↑) | 43.32% | 37.08% | 7.02% | 30.60% | 0% |
| | | | BR→TR(↑) | 23.19% | 19.97% | 3.39% | 14.95% | -8.20% |
| Punjabi | Gurumukhi | | Vanilla | 15.36 | 27.23 | 0.58 | 36.33 | 0.64 |
| | | | TR(Best) | 17.52 | 27.91 | 0.57 | 44.14 | 0.62 |
| | | | BR(Best) | 12.03 | 18.97 | 0.63 | 40.25 | 0.63 |
| | | | Vanilla→TR(↑) | 14.06% | 2.50% | -1.72% | 21.50% | -3.13% |
| | | | BR→TR(↑) | 45.64% | 47.13% | -9.52% | 9.66% | -1.59% |
| | | | Avg(Vanilla→TR(↑)) | **44.87%** | **20.75%** | -7.54% | **21.06%** | -8.92% |
| | | | Avg(BR→TR(↑)) | **25.75%** | **20.19%** | -10.17% | **10.58%** | -0.70% |
| Santali | Olchiki | | Vanilla | 0.34 | 0.63 | 0.62 | 18.79 | 0.35 |
| | | | TR(Best) | 9.49 | 12.24 | 0.67 | 20.71 | 0.41 |
| | | | BR(Best) | 13.12 | 16.51 | 0.63 | 20.18 | 0.42 |
| | | | Vanilla→TR(↑) | 2691.18% | 1842.86% | 8.06% | 10.22% | 17.14% |
| | | | BR→TR(↑) | -27.67% | -25.86% | 6.35% | 2.63% | -2.38% |
| Konkani | Devanagari | Low | Vanilla | 0.88 | 1.86 | 0.31 | 27.89 | 0.56 |
| | | | TR(Best) | 16.06 | 18.81 | 0.38 | 36.29 | 0.51 |
| | | | BR(Best) | 0.21 | 0.71 | 0.31 | 35.58 | 0.58 |
| | | | Vanilla→TR(↑) | 1725% | 911.29% | 22.58% | 30.12% | -8.93% |
| | | | BR→TR(↑) | 7547.62% | 2549.30% | 22.58% | 2% | -12.07% |
| Bodo | Devanagari | | Vanilla | 0.44 | 0.89 | 0.09 | 18.42 | 0.29 |
| | | | TR(Best) | 15.89 | 20.31 | 0.37 | 31.56 | 0.58 |
| | | | BR(Best) | 14.69 | 17.12 | 0.41 | 26.65 | 0.53 |
| | | | Vanilla→TR(↑) | 3511.36% | 2182.02% | 311.11% | 71.34% | 100% |
| | | | BR→TR(↑) | 8.17% | 18.63% | -9.76% | 18.42% | 9.43% |
| | | | Avg(Vanilla→TR(↑)) | **2642.51%** | **1645.39%** | **113.92%** | **37.23%** | **36.07%** |
| | | | Avg(BR→TR(↑)) | **2509.37%** | **847.36%** | **6.39%** | **7.68%** | -1.67% |

Table 1: Vanilla LLM's performance comparison after CPT with **TR=Top Rank**, **BR=Bottom Rank** small size (≤30K) corpus for various Indic languages covering different scripts and resource types. We report the performance improvement from **Vanilla→TR** and **BR→TR**. We also report the average improvement across resource type availability as **Avg(Vanilla→TR(↑))** and **Avg(BR→TR(↑))**, positive improvements are marked in **bold** and underlined.

### 4.3 Vocabulary augmentation helps in specific cases

In previous sections, we observed CPT with a small corpus improves LLM performance for most tasks and languages. To check if the performance can be improved further, we attempt vocabulary augmentation. Our hypothesis is that vocabulary augmentation would typically work for those languages where fragment ratio (average number of tokens generated per word) is high. We find the fragment ratio of the nine languages (Table 7) and group them into large, medium and small. We compare LLM performance with and without vocabulary augmentation while running CPT with **TR(Best)** and report the average improvement in Table 2. We experiment with addition of 100, 200 and 300 tokens and report the best result. We see vocabulary augmentation helps multi-word generation tasks like CrossSum and Flores(en→xx), when the fragmentation ratio is medium to large.

| Lang | Script | Fragment | CPT data | +Vocab | Chrf++ CrossSum | Chrf++ Flores(en→xx) | Token-F1 XorQA(xx) | Chrf++ Flores(xx→en) | Token-F1 XorQA(en) |
|---|---|---|---|---|---|---|---|---|---|
| Santali | OlChiki | Large | TR(Best) | No | 9.49 | 12.24 | 0.67 | 20.71 | 0.41 |
| | | | | Yes | 13.97 | 13.99 | 0.26 | 14.54 | 0.32 |
| | | | | chg(↑) | 47.21% | 14.30% | -61.19% | -29.79% | -21.95% |
| Telugu | Telugu | | TR(Best) | No | 16.51 | 25.57 | 0.37 | 39.31 | 0.59 |
| | | | | Yes | 18.13 | 26.03 | 0.36 | 43.59 | 0.61 |
| | | | | chg(↑) | 9.81% | 1.80% | -2.70% | 10.89% | 3.39% |
| | | | | Avg chg | **28.51%** | **8.05%** | -31.95% | -9.45% | -9.28% |
| Assamese | Bengali | Medium | TR(Best) | No | 15.78 | 21.81 | 0.61 | 39.52 | 0.56 |
| | | | | Yes | 16.63 | 21.92 | 0.54 | 38.57 | 0.64 |
| | | | | chg(↑) | 5.39% | 0.50% | -11.48% | -2.40% | 14.29% |
| Bengali | Bengali | | TR(Best) | No | 17.35 | 28.97 | 0.63 | 43.42 | 0.58 |
| | | | | Yes | 17.94 | 28.27 | 0.63 | 43.27 | 0.65 |
| | | | | chg(↑) | 3.40% | -2.42% | 0% | -0.35% | 12.07% |
| Punjabi | Gurumukhi | | TR(Best) | No | 17.52 | 27.91 | 0.57 | 44.14 | 0.62 |
| | | | | Yes | 17.34 | 28.44 | 0.56 | 39.73 | 0.59 |
| | | | | chg(↑) | -1.03% | 1.90% | -1.75% | -9.99% | -4.84% |
| | | | | Avg chg | **2.59%** | -0.01% | -4.41% | -4.25% | **7.17%** |
| Sanskrit | Devanagari | Small | TR(Best) | No | 13.63 | 15.15 | 0.31 | 33.71 | 0.42 |
| | | | | Yes | 13.84 | 14.14 | 0.36 | 28.31 | 0.41 |
| | | | | chg(↑) | 1.54% | -6.67% | 16.13% | -16.02% | -2.38% |
| Bodo | Devanagari | | TR(Best) | No | 15.89 | 20.31 | 0.37 | 31.56 | 0.58 |
| | | | | Yes | 17.12 | 20.51 | 0.49 | 30.11 | 0.51 |
| | | | | chg(↑) | 7.74% | 0.98% | 32.43% | -4.59% | -12.07% |
| Konkani | Devanagari | | TR(Best) | No | 16.06 | 18.81 | 0.38 | 36.29 | 0.51 |
| | | | | Yes | 15.12 | 15.95 | 0.46 | 31.52 | 0.36 |
| | | | | chg(↑) | -5.85% | -15.20% | 21.05% | -13.14% | -29.41% |
| Urdu | Arabic | | TR(Best) | No | 22.29 | 31.51 | 0.31 | 45.46 | 0.58 |
| | | | | Yes | 21.41 | 27.76 | 0.47 | 42.77 | 0.62 |
| | | | | chg(↑) | -3.95% | -11.90% | 51.61% | -5.92% | 6.90% |
| | | | | Avg chg | -0.13% | -8.20% | **30.31%** | -9.92% | -9.24% |

Table 2: Comparing LLM's performance w/o and w/ vocabulary augmentation (≤300) along with CPT with small size (≤30K) ranked training corpus for various Indic languages covering different scripts and resource types. We segregate the language (Large/Medium/Small) as per their fragmentation ratio reported in Table 7 and report individual and average performance changes across different levels of fragmentation, positive improvements are marked **bold** and underlined.

At lower levels of fragment ratios, we do not see benefits from vocabulary augmentation. In case of XorQA(xx), we see performance *drop* after vocabulary augmentation, for languages with a high fragment ratio. Poor initialization of the newly augmented words, followed by limited training, may hamper their single-word prediction abilities. We also discuss few error cases of XorQA(xx) in Appendix A.1.

For Flores(xx→en) and XorQA(en), where the target language is English, we do not see any improvement from vocabulary augmentation. This may be because we are adding Indic language-specific vocabulary and training with that language-specific corpus, giving no or negative improvement for English target tasks (we discuss it in Section 4.5). Another interesting observation is that with vocabulary augmentation, the LLM can generate more tokens than vanilla or without vocabulary-augmented LLM, given a similar output generation limit (more details on Appendix Section A.2).

### 4.4 Additional corpus and tokens not always helpful

To check if CPT with a larger corpus size and an order of magnitude large vocabulary size results in even better performance, we conduct CPT with 100K ranked corpus and 2000 additional vocabulary and compare it with 30K ranked corpus and 300 additional vocabulary. In Table 3, we report the result of these two configurations and find that a large CPT corpus with more additional vocabulary does not improve the performance as compared to a small-size corpus and vocabulary augmentation. This can be due to two reasons, first, as we are ranking the corpus, it might be possible most informative sentences are already present in the smaller 30K corpus. Second, as we are doing cost-efficient CPT by using LoRA and limited training steps (2 epochs), a large corpus with more additional vocabulary finds it difficult to converge, resulting in sub-optimal performance.

### 4.5 Adding English corpus to CPT improves English generation

In Table 2, we see LLM's performance drops for English target generation tasks like Flores(xx→en) and XorQA(en) after CPT using additional vocabulary. We hypothesize that this can be due to catastrophic forgetting as English corpus is not used while doing CPT. To verify this we add 20K randomly selected English sentence corpus with existing 30K Indic language-specific ranked cor-pus for CPT. In Table 4, we compare the LLM's performance after doing CPT with and without 20K English sentence corpus. We see that in almost all the cases, performance improves or remains the same as compared to CPT with only language-specific corpus. This justifies adding target language-specific corpus before CPT.

| | | | Metric→ | | Chrf++ | Chrf++ | Token-F1 | Chrf++ | Token-F1 |
|---|---|---|---|---|---|---|---|---|---|
| Lang | Script | Type | CPT data | +Vocab | CrossSum | Flores(en→xx) | XorQA(xx) | Flores(xx→en) | XorQA(en) |
| **Urdu** | Arabic | | 30K | 300 | 20.69 | 27.17 | 0.44 | 40.52 | 0.56 |
| | | | 100K | 2000 | 23.19 | 30.79 | 0.37 | 39.37 | 0.51 |
| | | | | chg | 12.08% | 13.32% | -15.91% | -2.84% | -8.93% |
| **Bengali** | Bengali | High | 30K | 300 | 17.32 | 27.61 | 0.63 | 37.02 | 0.61 |
| | | | 100K | 2000 | 19.29 | 29.67 | 0.49 | 39.07 | 0.55 |
| | | | | chg | 11.37% | 7.46% | -22.22% | 5.54% | -9.84% |
| **Telugu** | Telugu | | 30K | 300 | 18.13 | 24.19 | 0.31 | 41.88 | 0.61 |
| | | | 100K | 2000 | 18.16 | 26.13 | 0.17 | 31.59 | 0.54 |
| | | | | chg | 0.17% | 8.02% | -45.16% | -24.57% | -11.48% |
| | | | | Avg.chg | **<u>7.87%</u>** | **<u>9.6%</u>** | -27.76% | -7.29% | -10.08% |
| **Sanskrit** | Devanagari | | 30K | 300 | 12.02 | 12.98 | 0.37 | 26.39 | 0.47 |
| | | | 100K | 2000 | 9.06 | 13.91 | 0.22 | 25.44 | 0.41 |
| | | | | chg | -24.63% | 7.16% | -40.54% | -3.6% | -12.77% |
| **Assamese** | Bengali | Mid | 30K | 300 | 16.67 | 22.38 | 0.55 | 35.92 | 0.54 |
| | | | 100K | 2000 | 16.69 | 23.29 | 0.46 | 35.56 | 0.47 |
| | | | | chg | 0.12% | 4.07% | -16.36% | -1% | -12.96% |
| **Punjabi** | Gurumukhi | | 30K | 300 | 17.41 | 28.78 | 0.53 | 41.78 | 0.59 |
| | | | 100K | 2000 | 16.81 | 26.32 | 0.33 | 11.01 | 0.47 |
| | | | | chg | -3.45% | -8.55% | -37.74% | -73.65% | -20.34% |
| | | | | Avg.chg | -9.32% | **<u>0.89%</u>** | -31.55% | -26.08% | -15.36% |
| **Santali** | Ol Chiki | | 30K | 300 | 12.66 | 13.02 | 0.17 | 13.75 | 0.36 |
| | | | 100K | 2000 | 10.89 | 4.49 | 0.05 | 14.91 | 0.22 |
| | | | | chg | -13.98% | -65.51% | -70.59% | 8.44% | -38.89% |
| **Konkani** | Devanagari | Low | 30K | 300 | 15.45 | 15.81 | 0.37 | 30.96 | 0.31 |
| | | | 100K | 2000 | 15.51 | 20.15 | 0.38 | 30.13 | 0.34 |
| | | | | chg | 0.39% | 27.45% | 2.7% | -2.68% | 9.68% |
| **Bodo** | Devanagari | | 30K | 300 | 16.83 | 19.51 | 0.46 | 30.55 | 0.49 |
| | | | 100K | 2000 | 16.83 | 21.08 | 0.44 | 32.19 | 0.53 |
| | | | | chg | 0% | 8.05% | -4.35% | 5.37% | 8.16% |
| | | | | Avg.chg | -4.53% | -10% | -24.08% | **<u>3.71%</u>** | -7.02% |

Table 3: Comparing LLM's performance after CPT with 30K corpus, 300 additional vocabulary with 100K corpus with 2000 additional vocabulary for various Indic languages covering different scripts and resource types. Positive average improvements are marked **bold** and <u>underlined</u>.

| | | | Metric→ | Chrf++ | Token-F1 |
|---|---|---|---|---|---|
| Lang | CPT data | +Vocab | | Flores(xx→en) | XorQA(en) |
| **Urdu** | 30K | Yes | | 40.52 | 0.56 |
| | +20K(En) | Yes | | **<u>40.72</u>** | **<u>0.63</u>** |
| **Bengali** | 30K | Yes | | **<u>40.58</u>** | **<u>0.61</u>** |
| | +20K(En) | Yes | | 40.55 | 0.56 |
| **Telugu** | 30K | Yes | | 41.88 | 0.61 |
| | +20K(En) | Yes | | **<u>43.91</u>** | 0.61 |
| **Sanskrit** | 30K | Yes | | 26.39 | **<u>0.47</u>** |
| | +20K(En) | Yes | | **<u>28.01</u>** | 0.37 |
| **Assamese** | 30K | Yes | | 38.49 | **<u>0.61</u>** |
| | +20K(En) | Yes | | **<u>38.92</u>** | 0.58 |
| **Punjabi** | 30K | Yes | | 41.78 | 0.59 |
| | +20K(En) | Yes | | **<u>43.55</u>** | 0.59 |
| **Santali** | 30K | Yes | | 14.54 | 0.32 |
| | +20K(En) | Yes | | **<u>17.93</u>** | **<u>0.39</u>** |
| **Konkani** | 30K | Yes | | **<u>30.96</u>** | 0.31 |
| | +20K(En) | Yes | | 29.49 | **<u>0.45</u>** |
| **Bodo** | 30K | Yes | | 30.55 | 0.49 |
| | +20K(En) | Yes | | **<u>32.08</u>** | **<u>0.61</u>** |

Table 4: Comparing LLM's performance on English target generation tasks w/o and w/ additional 20K English corpus along with 30K ranked CPT corpus for various Indic languages. covering different scripts and resource types. Best performances are marked **bold** and <u>underlined</u>.

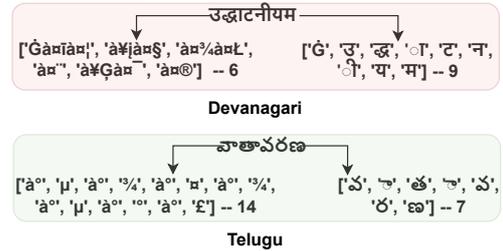

Figure 3: Number of tokens (mentioned in numbers) generated before and after the vocabulary augmentation for Devanagari and Telugu scripts. Red and Green shades indicate an increase and decrease of tokens, respectively, after vocabulary addition. (The strange-looking characters are not typesetting aberrations.)

### 4.6 Effect of tokenization after vocabulary addition

Finally, we study the LLM's tokenizer capability before and after adding additional vocabulary. In general, extra vocabulary can improve tokenization and generate a lesser number of tokens, which helps to reduce generation costs. In Table 5, we show the average number of tokens generated by the LLM's tokenizer before and after adding additional vocabulary for all languages and tasks. In our case, we stick to adding single-character tokens whenever possible, as this can help to transfer the CPT benefit to downstream tasks. During CPT, if we add multi-character tokens, it might be possible that the downstream tasks may not have that token, resulting in not passing the training benefit to target tasks. We observe that additional vocabulary augmentation reduces the average tokens per word except for the language having Devanagari scripts (Bodo, Konkani, and Sanskrit) and Arabic scripts (Urdu). In Figure 3, we show two examples of tokenization with and without additional vocabulary augmentation. As for Devanagari scripts, the word उद्घाटनीयम splits into 6 tokens, whereas, after vocabulary addition, it splits into 9 tokens. This is due to the fact that vanilla LLM tokenizer already splits the word better than single character split, but when we add single character tokens as additional vocabulary, it worsens the tokenization. However, for Telugu scripts where the word వాతావరణ splits into 14 tokens (single character splits into multiple bytes), single character token addition improves the tokenization by splitting it into 7 characters. As expected for English target generation tasks Flores(xx→en) and XorQA(en), no change is observed before and after the additional vocabulary augmentation, as we are only adding Indic language-specific tokens.

| Lang | Script | Type | CPT data | +Vocab | CrossSum | Flores(en→xx) | Flores(xx→en) | XorQA(xx) | XorQA(en) |
|---|---|---|---|---|---|---|---|---|---|
| Bengali | Bengali | High | Vanilla | - | **192.95** | **149.58** | 27.62 | **20.90** | 4.34 |
|  |  |  | 30K | 300 | 158.37 | 123.22 | 27.62 | 16.23 | 4.34 |
| Telugu | Telugu | High | Vanilla | - | **277.20** | **223.01** | 27.28 | **6.00** | 4.34 |
|  |  |  | 30K | 300 | 113.22 | 93.88 | 27.28 | 2.00 | 4.34 |
| Assamese | Bengali | Mid | Vanilla | - | **179.91** | **157.89** | 27.44 | **22.04** | 4.34 |
|  |  |  | 30K | 300 | 98.09 | 91.06 | 27.44 | 12.39 | 4.34 |
| Punjabi | Gurumukhi | Mid | Vanilla | - | **233.87** | **206.14** | 25.49 | **25.47** | 4.34 |
|  |  |  | 30K | 300 | 112.97 | 99.01 | 25.49 | 12.21 | 4.34 |
| Santali | Ol Chiki | Low | Vanilla | - | **353.90** | **344.09** | 25.58 | **40.85** | 4.34 |
|  |  |  | 30K | 300 | 142.88 | 138.85 | 25.58 | 15.50 | 4.34 |
| Konkani | Devanagari | Low | Vanilla | - | 82.70 | 72.73 | 27.60 | 10.44 | 4.34 |
|  |  |  | 30K | 300 | **110.09** | **98.55** | 27.60 | **13.96** | 4.34 |
| Bodo | Devanagari | Low | Vanilla | - | 86.51 | 83.86 | 27.63 | 8.85 | 4.34 |
|  |  |  | 30K | 300 | **98.32** | **92.41** | 27.60 | **10.90** | 4.34 |
| Sanskrit | Devanagari | Mid | Vanilla | - | 82.41 | 69.05 | 26.97 | 9.31 | 4.34 |
|  |  |  | 30K | 300 | **106.14** | **90.00** | 26.97 | **12.22** | 4.34 |
| Urdu | Arabic | High | Vanilla | - | 96.98 | 80.24 | 25.95 | 9.42 | 4.34 |
|  |  |  | 30K | 300 | **155.19** | **125.79** | 25.95 | **13.56** | 4.34 |

Table 5: Comparing the average number of tokens generated by the LLM before and after CPT with 30K and 300 additional vocabularies for all the tasks across Indic languages covering different scripts and resource types. The highest values are marked **bold** and underlined.

***Summary of observations***: Combining the observations from Tables 1, 2 and 5, we see all the languages benefit from CPT without vocabulary augmentation, though the degree of improvement is more for low resource languages. However, a similar pattern of improvement is absent when we augment additional vocabulary during CPT; here, we see improvement only if the language is over-fragmented by the LLM's tokenizer, irrespective of their resource availability type. As an example, although the language Bodo is resource-poor, it has a lesser fragment ratio (Table 7) as it shares the resource-rich Devanagari script, failing to reap the benefit of vocabulary augmentation. On the other hand, Santali, both a resource-poor and over-fragmented language (Table 7), get additional gain after vocabulary augmentation. So, our conclusion from this whole exercise is our method works best for a language which is poor in both terms, resources and script representation.

## 5 Conclusion

In this work, we proposed a technique based on LLM-sensitive and global scores to select a small-size CPT corpus. Further, we proposed a method to augment the LLM vocabulary with a limited number of new tokens. We showed the effectiveness of our selection methods, using IndicGen-Bench, an Indic language-specific generation task benchmark dataset, in experiments involving *nine* Indian languages covering different scripts and resource availability. Our results showed performance improvement after doing CPT with a small-sized corpus, and further improvement is possible in some situations, with limited vocabulary augmentation. In addition, we showed the improvements are not uniform across scripts, and more token addition may not necessarily result in performance gains. We found interesting observations related to tokenization across languages, and identified that language-specific CPT can hurt English generation performance. We believe that our work will be beneficial to future tasks involving LLMs for low-resource languages.

# 6 Limitations

Although recently, many white box LLMs like Llama families, Mistral, Phi, Gemma, etc., are available; we have only experimented with the LLama-3-8B model to work within our computation budget and carry out experiments with various languages and tasks. Though we stick to only one LLM for our research, including more LLMs in our study would be more insightful. To initialize the newly added word embedding, we use only the mean pooling method, which takes the average embedding of constituent tokens produced by the existing tokenizer. Although there are methods of embedding initialization like FOCUS, Merge, Align, Random, etc., we choose to mean as existing studies (Yamaguchi et al., 2024b; Tejaswi et al., 2024) show that it produces comparable results despite being simple. Having said that, considering other embedding techniques can make the study more comprehensive. Lastly, we restrict our experiment to only Indic languages and a few generation tasks; adding resource-poor languages from other language families and some more generation and classification tasks can strengthen our study further. We leave addressing these issues to future work.

# References

Orevaoghene Ahia, Sachin Kumar, Hila Gonen, Jungo Kasai, David R. Mortensen, Noah A. Smith, and Yulia Tsvetkov. 2023. Do all languages cost the same? tokenization in the era of commercial language models.

Abhinand Balachandran. 2023. Tamil-llama: A new tamil language model based on llama 2.

Yejin Bang, Samuel Cahyawijaya, Nayeon Lee, Wenliang Dai, Dan Su, Bryan Wilie, Holy Lovenia, Ziwei Ji, Tiezheng Yu, Willy Chung, Quyet V. Do, Yan Xu, and Pascale Fung. 2023. A multitask, multilingual, multimodal evaluation of chatgpt on reasoning, hallucination, and interactivity.

Ethan C. Chau, Lucy H. Lin, and Noah A. Smith. 2020. Parsing with multilingual BERT, a small corpus, and a small treebank. In *Findings of the Association for Computational Linguistics: EMNLP 2020*, pages 1324–1334, Online. Association for Computational Linguistics.

Aakanksha Chowdhery et al. 2022. Palm: Scaling language modeling with pathways.

Yiming Cui, Ziqing Yang, and Xin Yao. 2024. Efficient and effective text encoding for chinese llama and alpaca.

Konstantin Dobler and Gerard de Melo. 2023. FOCUS: Effective embedding initialization for monolingual specialization of multilingual models. In *Proceedings of the 2023 Conference on Empirical Methods in Natural Language Processing*, pages 13440–13454, Singapore. Association for Computational Linguistics.

C.m. Downey, Terra Blevins, Nora Goldfine, and Shane Steinert-Threlkeld. 2023. Embedding structure matters: Comparing methods to adapt multilingual vocabularies to new languages. In *Proceedings of the 3rd Workshop on Multi-lingual Representation Learning (MRL)*, pages 268–281, Singapore. Association for Computational Linguistics.

Abhimanyu Dubey et al. 2024. The llama 3 herd of models.

Kazuki Fujii, Taishi Nakamura, Mengsay Loem, Hiroki Iida, Masanari Ohi, Kakeru Hattori, Hirai Shota, Sakae Mizuki, Rio Yokota, and Naoaki Okazaki. 2024. Continual pre-training for cross-lingual llm adaptation: Enhancing japanese language capabilities.

Takuro Fujii, Koki Shibata, Atsuki Yamaguchi, Terufumi Morishita, and Yasuhiro Sogawa. 2023. How do different tokenizers perform on downstream tasks in scriptio continua languages?: A case study in Japanese. In *Proceedings of the 61st Annual Meeting of the Association for Computational Linguistics (Volume 4: Student Research Workshop)*, pages 39–49, Toronto, Canada. Association for Computational Linguistics.

Leonidas Gee, Andrea Zugarini, Leonardo Rigutini, and Paolo Torroni. 2022. Fast vocabulary transfer for language model compression. In *Proceedings of the 2022 Conference on Empirical Methods in Natural Language Processing: Industry Track*, pages 409–416, Abu Dhabi, UAE. Association for Computational Linguistics.

Amr Hendy, Mohamed Abdelrehim, Amr Sharaf, Vikas Raunak, Mohamed Gabr, Hitokazu Matsushita, Young Jin Kim, Mohamed Afify, and Hany Hassan Awadalla. 2023. How good are gpt models at machine translation? a comprehensive evaluation.

Valentin Hofmann, Hinrich Schütze, and Janet Pierrehumbert. 2022. An embarrassingly simple method to mitigate undesirable properties of pretrained language model tokenizers. In *Proceedings of the 60th Annual Meeting of the Association for Computational Linguistics*.

Wenxiang Jiao, Wenxuan Wang, Jen tse Huang, Xing Wang, Shuming Shi, and Zhaopeng Tu. 2023. Is chatgpt a good translator? yes with gpt-4 as the engine.

Taku Kudo. 2018. Subword regularization: Improving neural network translation models with multiple subword candidates.

Taku Kudo and John Richardson. 2018. Sentencepiece: A simple and language independent subword tokenizer and detokenizer for neural text processing.

Benjamin Minixhofer, Fabian Paischer, and Navid Rekabsaz. 2022. WECHSEL: Effective initialization of subword embeddings for cross-lingual transfer of monolingual language models. In *Proceedings of the 2022 Conference of the North American Chapter of the Association for Computational Linguistics: Human Language Technologies*, pages 3992–4006, Seattle, United States. Association for Computational Linguistics.

Benjamin Muller, Antonios Anastasopoulos, Benoît Sagot, and Djamé Seddah. 2021. When being unseen from mBERT is just the beginning: Handling new languages with multilingual language models. In *Proceedings of the 2021 Conference of the North American Chapter of the Association for Computational Linguistics: Human Language Technologies*, pages 448–462, Online. Association for Computational Linguistics.

Arijit Nag, Animesh Mukherjee, Niloy Ganguly, and Soumen Chakrabarti. 2024. Cost-performance optimization for processing low-resource language tasks using commercial llms.

OpenAI et al. 2023. Gpt-4 technical report.

Malte Ostendorff and Georg Rehm. 2023. Efficient language model training through cross-lingual and progressive transfer learning.

Lawrence Page, Sergey Brin, Rajeev Motwani, and Terry Winograd. 1999. The pagerank citation ranking : Bringing order to the web. In *The Web Conference*.

Aleksandar Petrov, Emanuele La Malfa, Philip H. S. Torr, and Adel Bibi. 2023. Language model tokenizers introduce unfairness between languages.

Maja Popović. 2017. chrF++: words helping character n-grams. In *Proceedings of the Second Conference on Machine Translation*, pages 612–618, Copenhagen, Denmark. Association for Computational Linguistics.

Phillip Rust, Jonas Pfeiffer, Ivan Vulić, Sebastian Ruder, and Iryna Gurevych. 2021. How good is your tokenizer? on the monolingual performance of multilingual language models. In *Proceedings of the 59th Annual Meeting of the Association for Computational Linguistics and the 11th International Joint Conference on Natural Language Processing (Volume 1: Long Papers)*, pages 3118–3135, Online. Association for Computational Linguistics.

Mike Schuster and Kaisuke Nakajima. 2012. Japanese and korean voice search. In *2012 IEEE International Conference on Acoustics, Speech and Signal Processing (ICASSP)*, pages 5149–5152.

Rico Sennrich, Barry Haddow, and Alexandra Birch. 2016. Neural machine translation of rare words with subword units.

Harman Singh, Nitish Gupta, Shikhar Bharadwaj, Dinesh Tewari, and Partha Talukdar. 2024. Indicgenbench: A multilingual benchmark to evaluate generation capabilities of llms on indic languages.

Atula Tejaswi, Nilesh Gupta, and Eunsol Choi. 2024. Exploring design choices for building language-specific llms.

Cagri Toraman, Eyup Halit Yilmaz, Furkan Şahinuç, and Oguzhan Ozcelik. 2023. Impact of tokenization on language models: An analysis for turkish. *ACM Trans. Asian Low-Resour. Lang. Inf. Process.*, 22(4).

Hugo Touvron et al. 2023. LLaMA 2: Open foundation and fine-tuned chat models.

Atsuki Yamaguchi, Aline Villavicencio, and Nikolaos Aletras. 2024a. An empirical study on cross-lingual vocabulary adaptation for efficient language model inference.

Atsuki Yamaguchi, Aline Villavicencio, and Nikolaos Aletras. 2024b. How can we effectively expand the vocabulary of llms with 0.01gb of target language text?

# Efficient Continual Pre-training of LLMs for Low-resource Languages
(Appendix)

## A Supplementary results

| | |
|---|---|
| CrossSum (Summarization) | Text: By Leo KelionTechnology desk editor The alleged cyber-weapons are said to include malware that targets Windows, Android, iOS, OSX and Linux computers as well as internet routers. Some of the software is "Reported [...TRUNCATED...]. <br><br> Summary(Assamese): Wikileaksএ এনে কিছু সবিশেষ প্ৰকাশ কৰিছে, যিবোৰ ইয়াৰ মতে এইবোৰ হৈছে চিআইএৰ দ্বাৰা ব্যৱহৃত বিস্তৃত পৰিসৰৰ হেকিং সঁজুলি। |
| Flores(en-xx) (Machine Translation) | Source: The Luno had 120–160 cubic metres of fuel aboard when it broke down and high winds and waves pushed it into the breakwater. <br><br> Target(Assamese): লুন' যেতিয়া ধ্বংসপ্ৰাপ্ত হৈছিল আৰু তীব্ৰ বতাহ আৰু ঢৌৱে ইয়াক ঠেলি নি পাৰৰ বান্ধবোৰত খুন্দিয়াইছিল, তেতিয়া তাত 120–160 বৰ্গমিটাৰ ইন্ধন মজুত আছিল। |
| Flores(xx-en) (Machine Translation) | Source(Assamese): লুন' যেতিয়া ধ্বংসপ্ৰাপ্ত হৈছিল আৰু তীব্ৰ বতাহ আৰু ঢৌৱে ইয়াক ঠেলি নি পাৰৰ বান্ধবোৰত খুন্দিয়াইছিল, তেতিয়া তাত 120–160 বৰ্গমিটাৰ ইন্ধন মজুত আছিল। <br><br> Target: The Luno had 120–160 cubic metres of fuel aboard when it broke down and high winds and waves pushed it into the breakwater. |
| XorQA(xx) (Question-Answering) | Context: Al-Mansur was born at the home of the Abbasid family in Humeima (modern-day Jordan) after their emigration from the Hejaz in 95 AH (714 CE). His father, Muhammad, was reputedly a great grandson of Abbas ibn Abd al-Muttalib, the youngest uncle of Mohammad. His mother, as described in the 14th-century Moroccan historical work Rawd al-Qirtas, was one Sallama [...TRUNCATED...] <br><br> Question(Assamese): দ্বিতীয় আব্বাচীদ খলিফা আবু জাফৰ আব্দুল্লাহ বিন মুহাম্মাদ আল মনচুৰৰ মাতৃৰ নাম কি ? <br><br> Answer(Assamese): চাল্লামা |
| XorQA(en) (Question-Answering) | Context: Al-Mansur was born at the home of the Abbasid family in Humeima (modern-day Jordan) after their emigration from the Hejaz in 95 AH (714 CE). His father, Muhammad, was reputedly a great grandson of Abbas ibn Abd al-Muttalib, the youngest uncle of Mohammad. His mother, as described in the 14th-century Moroccan historical work Rawd al-Qirtas, was one Sallama [...TRUNCATED...] <br><br> Question(Assamese): দ্বিতীয় আব্বাচীদ খলিফা আবু জাফৰ আব্দুল্লাহ বিন মুহাম্মাদ আল মনচুৰৰ মাতৃৰ নাম কি ? <br><br> Answer: Sallama |

Figure 4: Example instance from each dataset.

### A.1 Error cases for QA tasks

We show a few error cases in Table 6, where the LLM fails after vocabulary augmentation for XorQA(xx). In one such case, the prediction is correct after vocabulary augmentation, but the evaluation metric flags it as incorrect owing to different wording. E.g., ৮৭,০০০ and ৮৭ হাজাৰ have same meaning as in Assamese হাজাৰ means 1000. There are cases where we find the vocabulary-augmented LM generates the correct response, but in English. Also, there are cases where the LM stopped generation after producing the first character, which is correct. This can be due to the adverse effect of change in vocabulary distribution after augmentation. Another case is possibly related to the undesirable change in vocabulary distribution where the model starts with newly added tokens and ultimately produces the wrong outcome.

| Cases | Gold label | w/o Vocab add | w/ Vocab add |
|---|---|---|---|
| **Correct but different wording** | ৮৭,০০০ (87,000) | ৮৭,০০০ | ৮৭ হাজাৰ (হাজাৰ = 1000) |
| **Correct but in English** | ৩৫ (35) | ৩৫ | 35 |
| **Stopped after few character** | నాసా (NASA) | నాస | నా |
| **Started with added vocab and failed** | পশ্চিম বংগ (West Bengal) | পশ্চিম বংগ | বি বংগা. |

Table 6: Error cases for XorQA(xx) tasks. The second, third and fourth columns show the gold label, predication without and with additional vocabulary augmentation, respectively, for a particular question given a context. Important information related to the answers are underlined.

### A.2 Vocabulary augmentation helps generate more tokens

An interesting observation is with vocabulary augmentation, the LLM can generate more tokens than vanilla or without vocabulary-augmented LM, given a similar output generation limit. Consequently, sometimes it can improve summarization performance by extracting more information about the context. As shown in Figure 6, the LLM generates a longer and more informative summary of the given paragraph after vocabulary augmentation. However, a thorough investigation is needed to check if more generations are always linked with more relevant information.

### A.3 Fragment ratio

| Language | Fragment ratio |
|---|---|
| Santali | 13.67 |
| Telugu | 12.44 |
| Assamese | 8.82 |
| Bengali | 8.04 |
| Punjabi | 7.54 |
| Sanskrit | 5.32 |
| Bodo | 3.89 |
| Konkani | 3.67 |
| Urdu | 2.85 |

Table 7: Degree of fragmentation on 30K rank training corpus for 9 Indic languages using LLama-3-8b model tokenizer.

## B Experimental settings

| Hyperparameter | Value |
|---|---|
| LLM | LLama-3 |
| LLM parameter size | 8 Billion |
| LLM model type | 8B-Instruct |
| LLM temperature | 0.5 (for summarization), 0.3(for translation), 0.001(for QA tasks) |
| LLM top p | 0.95 |
| Seed | 42 |
| LoRA r | 8 |
| LoRA alpha | 32 |
| LoRA dropout | 0.05 |
| LoRA task type | CAUSAL_LM |
| Learning rate | 1e-4 |
| Batch size | 32 |
| Epoch | 2 |
| $\alpha, \beta$ in Algorithm 1 & 2 | 0.5,0.5 |

Table 8: Details of LLM LoRA training and zero-shot inference hyperparameters.

We use meta-llama/MetaLlama38BInstruct model for our CPT and zero-shot inferencing. We run all the experiments in a single 80GB A100 GPU system. To preserve cost, we do all the experiments one time, and to make them reproducible, we fix the seed value to 42. To run CPT with 30K data, it took around 3 hours on a single 80GB A100 GPU. For Zero-shot testing, for each task, we select 100 random instances for each language.

| Task | Prompt |
|---|---|
| CrossSum (Summarization) | **Prompt:** Summarize the following article in <desired Indic language>. Summary should be strictly within <gold summary word count> word limit.<br><br>**Article:** By Leo KelionTechnology desk editor The alleged cyber-weapons are said to include malware that targets Windows, Android, iOS, OSX and Linux computers as well as internet routers. Some of the software is "Reported […TRUNCATED…].<br><br>**Summary:** |
| Flores(en-xx) (Machine Translation) | **Prompt:** Translate the following source sentence to <desired Indic language>. Translation should be strictly within <gold translation word count> word limit.<br><br>**Source sentence:** The Luno had 120–160 cubic metres of fuel aboard when it broke down and high winds and waves pushed it into the breakwater.<br><br>**Translation:** |
| Flores(xx-en) (Machine Translation) | **Prompt:** Translate the following source sentence in <desired Indic language> to English. Translation should be strictly within <gold translation word count> word limit.<br><br>**Source sentence:** লুন' যেতিয়া ধ্বংসপ্ৰাপ্ত হৈছিল আৰু তীব্ৰ বতাহ আৰু ঢৌৱে ইয়াক ঠেলি নি পাৱাৰ বাঙ্কেৰোৱত খুন্দিয়াইছিল, তেতিয়া তাত ১২০–১৬০ বৰ্গমিটাৰ ইন্ধন মজুত আছিল।<br><br>**Translation:** |
| XorQA(xx) (Question-Answering) | **Prompt:** Answer the question from the given context. Answer should be strictly in <desired Indic language> and within <gold answer word count> word limit. Output only the answer.<br><br>**Context:** Al-Mansur was born at the home of the Abbasid family in Humeima (modern-day Jordan) after their emigration from the Hejaz in 95 AH (714 CE). His father, Muhammad, was reputedly a great-grandson of Abbas ibn Abd al-Muttalib, the youngest uncle of Mohammad. His mother, as described in the 14th-century Moroccan historical work Rawd al-Qirtas, was one Sallama […TRUNCATED…].<br><br>**Question:** দ্বিতীয় আব্বাচীদ খলিফা আবু জাফৰ আব্দুল্লাহ বিন মুহাম্মাদ আল মনচুৰৰ মাতৃৰ নাম কি?<br><br>**Answer:** |
| XorQA(en) (Question-Answering) | **Prompt:** Answer the question from the given context. Answer should be strictly in English and within <gold answer word count> word limit. Output only the answer.<br><br>**Context:** Al-Mansur was born at the home of the Abbasid family in Humeima (modern-day Jordan) after their emigration from the Hejaz in 95 AH (714 CE). His father, Muhammad, was reputedly a great-grandson of Abbas ibn Abd al-Muttalib, the youngest uncle of Mohammad. His mother, as described in the 14th-century Moroccan historical work Rawd al-Qirtas, was one Sallama […TRUNCATED…].<br><br>**Question:** দ্বিতীয় আব্বাচীদ খলিফা আবু জাফৰ আব্দুল্লাহ বিন মুহাম্মাদ আল মনচুৰৰ মাতৃৰ নাম কি?<br><br>**Answer:** |

Figure 5: Details of the prompts used for each task.

| | |
|---|---|
| Original context: | তিনি শুধু একজন খেলোয়াড়ই নন, একজন বিখ্যাত অধিনায়কও ছিলেন। তিনি তার ক্রিকেট জীবনে সর্বমোট ৩১১টি একদিনের আন্তর্জাতিক ম্যাচ খেলেছেন এবং ১১,৩৬৩ রান সংগ্রহ করেছেন। পাশাপাশি তিনি ১১৩টি টেস্ট খেলেছেন ও ৭,২১২ রান সংগ্রহ করেছেন। ভারতকে তিনি ৪৯টি টেস্ট ম্যাচে নেতৃত্ব দিয়েছিলেন যার মধ্যে ভারত জিতেছিল ২১টি ম্যাচে। [...] |
| Summary(w/o extra vocabulary): | তিনি একজন বিখ্যাত অধিনায়ক ও খেলোয়াড় ছিলেন। তিনি ৩১১টি একদিনের আন্তর্জাতিক ম্যাচ খেলেছেন এবং ১১,৩৬৩ রান সংগ্রহ করেছেন। পাশাপাশি তিনি ১১৩টি টেস্ট খেলেছেন ও ৭,২১২ রান সংগ্রহ করেছেন। তিনি ভারতকে ৪৯টি টেস্ট |
| Summary(w/ extra vocabulary): | তিনি একজন বিখ্যাত অধিনায়কও ছিলেন। তিনি ৩১১টি একদিনের আন্তর্জাতিক ম্যাচ খেলেছেন এবং ১১,৩৬৩ রান সংগ্রহ করেছেন। তিনি ৪৯টি টেস্ট খেলেছেন ও ৭,২১২ রান সংগ্রহ করেছেন। তিনি ভারতকে ১৪৬টি একদিনের আন্তর্জাতিক ম্যাচে নেতৃত্ব দিয়েছিলেন। তিনি ভারতের একজন মিডিয়াম পেসার বোলারও ছিলেন। তিনি একদিনের আন্তর্জাতিক ক্রিকেটে ১০০টি ও টেস্টে ৩২টি উইকেট দখল করেন। তিনি একদিনের আন্তর্জাতিকে ১০০টি ও টেস্টে ৭১টি ক্যাচ নিয়েছেন। তিনি ২০০৮ সালে অস্ট্রেলিয়ার |

Figure 6: Added vocabulary can help LLM generate more text compared to vanilla LLM, given the same output generation limit. Here the summaries generated w/ and w/o additional vocabulary augmentation are shown in Green and Red, respectively. We see that the summary generated w/ additional vocabulary contains more words and information compared to w/o extra vocabulary augmented model.